%% file: main.tex
\documentclass{article}

\usepackage{microtype}
\usepackage{graphicx}
\usepackage{subfigure}
\usepackage{booktabs} 
\usepackage{amssymb,amsfonts}

\usepackage{hyperref}

\usepackage[accepted]{icmlw2019generalization}

\icmltitlerunning{The Principle of Unchanged Optimality in Reinforcement Learning Generalization}

\newcommand{\E}[2]{\mathbb{E}_{#1}\left[#2\right]}

\usepackage{titlesec}
\titlespacing*{\section}{0pt}{0.0ex}{0.0ex}

\begin{document}

\twocolumn[
\icmltitle{The Principle of Unchanged Optimality in Reinforcement Learning Generalization}



\icmlsetsymbol{equal}{*}

\begin{icmlauthorlist}
\icmlauthor{Alex Irpan}{equal,goo}
\icmlauthor{Xingyou Song}{equal,goo}
\end{icmlauthorlist}

\icmlaffiliation{goo}{Google Brain}

\icmlcorrespondingauthor{Alex Irpan}{alexirpan@google.com}

\icmlkeywords{Machine Learning, ICML}
\vskip 0.3cm
]



\printAffiliationsAndNotice{\icmlEqualContribution} 

\begin{abstract}
Several recent papers have examined generalization in reinforcement learning (RL), by proposing new environments or ways to add noise to existing environments, then benchmarking algorithms and model architectures on those environments. We discuss subtle conceptual properties of RL benchmarks that are not required in supervised learning (SL), and also properties that an RL benchmark should possess. Chief among them is one we call \textit{the principle of unchanged optimality}: there should exist a single $\pi$ that is optimal across all train and test tasks. In this work, we argue why this principle is important, and ways it can be broken or satisfied due to subtle choices in state representation or model architecture. We conclude by discussing challenges and future lines of research in theoretically analyzing generalization benchmarks.
\end{abstract}
\fontdimen2\font=2.3pt
\section{Introduction}
\label{sect:intro}
Reinforcement learning has achieved strong results in many game-like settings, but has been criticized for its sample inefficiency and its capability to overfit to environments it was trained on. In order to test generalization, benchmarks on Atari and real world games ~\cite{nichol2018gotta,ale} inherently assume that there is some similarity in distributions between train and test, much like the assumption that benchmarks such as MNIST and CIFAR10 are sampled from a underlying distribution. 

To spur research into generalization research with more control over the environment variables, many recent papers~\citep{cobbe2018quantifying,zhang2018natural} have proposed ways to construct training and test environments, then evaluating approaches similarly to supervised learning: train in the train MDPs and report episode reward in the test MDPs. These benchmarks are usually designed by intuition. Artificially generated benchmarks assume that multiple parts of the MDP are parametrizable, such as the transition, initial state, and reward - this is reasonable, but does not come with a guarantee that the generalization problem is theoretically well-founded. 

\section{Generalization Problems in Reinforcement Learning}
It is important that agents produced by reinforcement learning successfully generalize. However, what ``generalize'' formally means is somewhat unclear. We present a formalism of generalization based on generalization in supervised learning.

\subsection{Supervised Learning and Reinforcement Learning's Generalization Gap}
\label{sect:supervised}
We briefly touch upon the definition of generalization in SL and produce a standard definition of the RL generalization gap, as formalized in other RL benchmarks, such as ~\cite{zhang2018dissection, reparametrizable}. 

In a supervised learning problem, there exists a ground truth data distribution $(x_{i}, y_{i}) \sim \mathcal{D}$, where $(x_{i}, y_{i}) \in \mathcal{X}, \mathcal{Y}$, are from the input set and label sets, respectively.

Within statistical learning theory, these generalization error sources can be separated into out-of-sample errors and out-of-distribution errors. A classifier is a function $f: \mathcal{X} \rightarrow \mathcal{Y}$, and given a loss $\ell(f(x), y)$, let the true error be $err = \E{(x,y) \sim \mathcal{D}}{\ell(f(x),y)}$ be the expected loss over $X \times Y$. The empirical sample error of $n$ points $(x_i,y_i)$ is $\hat{err}_{n} = \frac{1}{n} \sum_{i=1}^{n} \ell(f(x_{i}), y_{i}) $. The out-out-sample error is defined as the expected difference $|\hat{err}_{n} - err |$ over the randomness of sampling the $n$ points. Out-of-distribution error occurs when taking the difference on performance on another distribution's dataset.

In this work, we will focus on out-of-sample rather than out-of-distribution error, assuming train MDPs and test MDPs are from the same distribution.

\subsection{RL Notation}
Reinforcement learning aims to learn a policy $\pi$ that maximizes expected episode reward $R$ in a Markov Decision Process (MDP) $\mathcal{M}$. An MDP $\mathcal{M}$ is represented by the tuple $\mathcal{M} = (\mathcal{S},\mathcal{A},r,\mathcal{T},S_0)$, where $\mathcal{S}$ is the state space, $\mathcal{A}$ is the action space, $r(s,a)$ is the reward function, $\mathcal{T}(s,a) \mapsto s'$ is the dynamics function, and $S_0$ is the start state distribution. Conditioned on state $s$, policies $\pi(a|s)$ are distributions over $\mathcal{A}$. Rewards are discounted by a discount factor $\gamma \in [0,1]$. Letting $r_t = r(s_t,a_t)$ to simplify notation, episode reward $R(\pi)$ is defined as $\E{\pi}{\sum_{t=0}^\infty \gamma^tr_t}$. While $\gamma < 1$ is useful for a surrogate training objective, it is common to let $\gamma = 1$ when quantifying actual performance.

For this paper, we assume we are studying on-policy or near on-policy reinforcement learning, where the policy continually collects new experience in the environment, and does so often enough that training data is never too far from on-policy data. RL algorithms can train directly on new experience (on-policy), or on a replay buffer of recent experience (near on-policy).

\subsection{Creating Out-of-Sample RL Problems} \label{sec:out-of-sample-rl} 
We formalize the procedural generation process described in Section~\ref{sect:intro}. Let $\Theta$ be a distribution over parameters $\theta$ that parametrize an MDP family $\{\mathcal{M}_\theta : \theta \in \Theta\}$. Each $\theta$ encompasses some transformation or modification to reward, transitions, starting state, and observation function, with $\mathcal{M}_\theta = (\mathcal{S}_\theta, \mathcal{A}_\theta, r_\theta, \mathcal{T}_\theta, S_{0,\theta})$. An appropriate train, validation, and test set can then be created by randomly sampling $\theta \sim \Theta$ and training or evaluating within $\mathcal{M}_\theta$, and thus episodic reward will also be parametrized as $R_{\theta}(\pi)$. 

Let $\widehat{\Theta}_{n,train}$ be an empirical sample of $n$ parameters $\theta$, and suppose we train $\pi$ to optimize reward against $\{\mathcal{M}_\theta : \theta \sim \widehat{\Theta}_{n,train}\}$. The objective $J(\pi)$ maximized is the average reward over this empirical sample.
\[
J_{\widehat{\Theta}}(\pi) = \frac{1}{n} \sum_{\theta_{i} \in \widehat{\Theta}_{n, train} }{R_{\theta_{i}}(\pi) }
\]
We want to generalize to all $\theta \in \Theta$, which can be expressed as the average episode reward $R$ over the full distribution.
\[
    J_{\Theta}(\pi) = \E{\theta \sim \Theta}{R_{\theta}(\pi)}
\]

Thus it follows to define the generalization gap in RL as the expectation of $$ J_{\widehat{\Theta}}(\pi) -  J_{\Theta}(\pi) $$
This is a natural translation of the supervised learning paradigm to the reinforcement learning use-case. However, not all parametrizations of $\theta$ naturally lend themselves to useful reinforcement learning benchmarks.

\subsection{Principle of Unchanged Optimality}
A common trick in image-based classification is to add data augmentation to the image to transform the image $x$ into a new image $x'$, without changing the class label $y$. Common choices are to apply random reflections, or to randomize brightness and contrast. Some data augmentation methods like mixup~\citep{zhang2018mixup} apply transformations that change the class label $y$ as well, and although these approaches work well in practice, we ignore them for this paper due to lack of theoretical justification.

Given a picture of a dog or cat, we know it is still identifiable as a dog or cat after data augmentation, because our human perception can successfully identify the class of these augmented pictures. Similarly, overlaying different random textures as done in CoinRun \cite{cobbe2018quantifying} and other domain randomization setups does not change the action of the optimal policy.

This suggests an informal principle: \textbf{when designing a generalization benchmark, there should exist a $\pi^*$ which is optimal for all MDPs in the training, validation, and test sets}.

We take this principle for granted in supervised learning - since the loss value is fixed as $\ell(f(x_{i}), y_{i})$, the function $f$ simply needs to output the minimizer $\hat{y}_{i}$ of $\ell(\hat{y}_{i}, y_{i})$, and therefore we may let $f(x_{i}) = \hat{y}_{i}$ on this instance, with no problems on conflicting outputs - e.g. a picture cannot possess ground truth label as a dog \textit{and} a cat. Thus, for any dataset $(x,y) \sim \mathcal{D}$ and any loss function $\ell$, there always exists a function $f$ in the underlying \textit{function space} which minimizes both each \textit{individual loss} and the global average loss, and the practical problem is to find a model that can replicate this function. In RL, there may not exist such a policy in the underlying \textit{policy space} and parametrizing the policy space with a neural network will also have this problem.

One example where the principle holds true is the practice of taking an MDP $\mathcal{M}$ where all randomness is guided by a pseudorandom RNG, letting $\Theta$ be the set of all possible random seeds, $\theta \in \Theta$ be a specific seed, and sampling different seeds for train and test MDPs. This is the approach taken by~\citep{zhang2018dissection}, which studies MuJoCo MDPs $\mathcal{M} = (\mathcal{S}, \mathcal{A}, r, \mathcal{T}, S_0)$ where everything is deterministic except for the initial state distribution $S_0$. 
By treating each different random seed as a separate environment, we create a population of deterministic MDPs $\{(\mathcal{S}, \mathcal{A}, r, \mathcal{T}, \delta(s_{0,\theta}))\}$, where $\delta(s_{0,\theta})$ is the Dirac-delta distribution for initial state $s_0$ sampled when the random seed is set to $\theta$. By construction, the optimal $\pi^*$ in the original MDP $(\mathcal{S}, \mathcal{A}, r, \mathcal{T}, S_0)$ must be optimal for each individual deterministic MDP $\mathcal{M}_\theta$.

However, what is considered "random" for \textit{intra-MDP} (randomness occurring in one MDP) or \textit{inter-MDP} (randomness that makes two MDP's different) may be subjective. This can make questions about ``memorization'' and ``generalization'' tricky, since they depend on the point of view. For instance, one layer of "memorization" is taking a single MDP, unwrapping the random seeds to create a family of deterministic MDPs, learning a policy on a limited number of seeds, and showing it fails at all other seeds. A higher layer of memorization is when the policy performing well on any seed from a single stochastic MDP $\mathcal{M}_\theta$, but fails on other similar MDPs. These are both valid generalization problems. The care comes in carefully examining what kinds of algorithms are sufficient to successfully learn agents that generalize according to the generalization problem the benchmark implicitly defines, and whether a benchmarking testing for this kind of generalization is useful.

\section{Breaking the Unchanged Optimality Principle}
Suppose we modify the \textit{dynamics} of the environment, in a way that is not visible or exposed to the agent. 

As an example, consider tasks on the common MuJoCo suite. Let $\theta$ represent changes to the gravity, friction, time between simulator steps, or other aspects of the physics simulator. This affects transitions $\mathcal{T}_{\theta}(s,a)$, which affects which trajectories are feasible and which are not, which affects final optimal behavior.

In many MuJoCo tasks, the observation $s$ contains only the positions and velocities of each robot joint, and without further modification it is not possible to recover $\theta$, since changes in acceleration are not observable from velocity at a single time $t$. Therefore, a feedforward policy cannot simultaneously produce the optimal action $a$ for both low gravity and high gravity environments, because the generalization benchmark has turned the MDP into a POMDP where $\theta$ is not observed, and the optimal actions $a$ are genuinely different.

The meta-learning framework \cite{maml} attempts to solve this issue by replacing the objective with policy $\pi_{\theta}$ into the objective using gradient adapted policy $\pi_{\theta - \eta \nabla_{\theta}}$. This inherently changes the objective function, but this change may provide the existence of a optimal policy $\pi^{*}$ quickly adaptable to any $\theta \in \hat{\Theta}$ and $\theta \in \Theta$~\citep{finn2018metalearning}.

Note that if we define the learning objective as maximizing global average reward over the population of MDPs, it will have an optimum $\pi^*$, even if $\theta$ is not directly observed, but this $\pi^*$ will not \textit{also} be optimal for each individual MDP. This may be sufficient in practice if the aim is to learn a robust policy, rather than an individually optimal one~\cite{rarl,rajeswaran2017epopt}. However, doing so has problematic theoretical consequences.

\subsection{Consequences of Breaking Unchanged Optimality}
Suppose in our setting, each sampled MDP $\mathcal{M}_\theta$ has a different optimal $\pi^*$. Let $\pi^*_{train}$ be a global optima of the empirical reward $J_{\widehat{\Theta}}(\pi)$. How does $J_{\widehat{\Theta}}(\pi)$ compare to true reward $J_{\Theta}(\pi)$? 

The learning process first converges at the aligned optimums of a few of the $\mathcal{M}_{\theta}$, ignoring the other sampled MDP's. However, the learning process is pulled towards the average of each of these $\pi^*$ as $n \rightarrow \infty$ which produces poor performance. The sharp drop in generalization gap as $n$ increases is generally known as the memorization limit \cite{zhang2017understanding} and depends on the smoothness and convexities of the loss landscapes from the neural network parametrization of the policy space. 

This is a problem for theoretical analysis, since many of these analyses assumes the optimization procedure discovers some optimum for the optimization problem and proceeds from there. When the unchanged optimality principle is broken, the theoretically optimal policy for the training set no longer matches the final desired policy, and thus becomes more like a memorization benchmark. The fact that ignoring this principle works in practice is more of an empirical bet about the limitations of our optimization procedures and regularity between different $\mathcal{M}_\theta$, rather than any theoretical validation.

We emphasize that even when unchanged optimality is upheld, not all maximizers of empirical average reward $J_{\hat{\Theta}}(\pi)$ will be maximizers of the final objective $J_{\Theta}(\pi)$. However, we are guaranteed that there exists a $\pi^*$ that maximizes both \textit{individual reward} and \textit{global average reward} at once, and the practical problem is to explore algorithms and architectures that are more likely to learn said $\pi^*$.

\section{Recovering Unchanged Optimality}
\subsection{State Augmentation}
\label{sect:state-aug}
Recalling that we would like there to be a single $\pi^*$ that is optimal across all MDPs at once, a conceptual but impractical option is to augment the state-space with the sampled parameter $\theta$ itself, with $s' = concat(s, \theta)$. This unifies all transitions $\mathcal{T}_{\theta}$ into a single transition $\mathcal{T}_{universal}$ operating on this new space (i.e. the transition function can simply look at $\theta$ and pick which sub-transition to use). The optimality condition is satisfied because a policy can look at $\theta$ and pick the optimal subpolicy. However, there are many other example ways to perform this re-parametrization as well, such as concatenating all MDP spaces into one single long-vector space.~\cite{reparametrizable} also defines "reparametrizable" MDP's, where the reward function is unified, and the transition function $\mathcal{T}$ becomes deterministic after factoring out its randomness (which may come from different distributions) $\xi_{1},..., \xi_{T}$, which leads to theoretical analysis.

However, constructing theoretical analyses to encompass a wide variety other reparametrizations can be difficult. For example, consider the state augmentation from Section~\ref{sect:state-aug}. None of this augmentation changed the underlying RL algorithm. They only changed the policy representation. Some policy $\pi(a_t|s_t)$ is transformed to either $\pi(a_t|s_t,\theta)$, $\pi(a_t|s_1, a_1, \cdots, s_t)$, or $\pi(a_t|s_t,s_{t-1},s_{t-2},s_{t-3})$. It is difficult to compare across approaches, since each representation leads to a different set of potential policies $\pi$. At the same time, if $\pi$ is restricted to feed-forward policies $\pi(a_t|s_t)$, it is difficult to construct a \textit{mathematically provable} MDP family that meets the principle of unchanged optimality unless we assume some universal transition and reward functions $\mathcal{T}$ and $r(\cdot)$.

All of these methods touch on an informal rule: \textbf{when designing a generalization benchmark, $\pi$ should be given all necessary information to predict optimal action $a$.} As shown above, directly appending $\theta$ to observation $s_t$ is one way to do so. \citep{yu2017preparing} applied this approach, training a universal policy $\pi(a|x,\theta)$ with ground truth $\theta$, then training an online system identification model $\phi$ that regressed state-action history to an estimate $\hat{\theta}$ used at test time. Another approach is to incorporate state-action history directly into $\pi$, letting the model implicitly recover $\theta$. \citep{peng2018sim2real} combined dynamics randomization with an LSTM policy, and found the learned policy encoded $\theta$ within its internal state, allowing the model to implicitly perform online system identification. \cite{finn2018metalearning} establishes that an RNN can naturally perform gradient-like adaptation without explicitly being told to do so, and thus "adapting to $\theta$" is similar to meta-learning. DQN approaches on Atari usually stack the previous 4 frames~\citep{mnih2015human}, allowing the model to observe a truncated state history that can be sufficient for some games.

\section{Model-Based RL for Generalization}
Generalization requires partially learning $\theta$ to learn the dynamics. This naturally leads to a discussion of model-based RL, since these approaches aim to learn the dynamics. The model-based framework is designed to assist a policy in gaining information about the environment to improve sample efficiency for training \cite{modelbased, i2a, algoframework, merlin}. However, there is little work in conceptually modifying the model-based RL paradigm for generalization.

Recall that we stated that a generalization benchmark needs to provide information that helps in predicting the optimal action. This notion of ``necessary information'' is important for model learning as well. The learned dynamics model should provide or structure the information needed to predict the optimal action. If said dynamics model only provides necessary information for one or a small subset of the MDP family, the model will generalize poorly. However, such a model may help the policy train faster on this small subset, and the policy could rely on this non-generalizable information.

Taking video prediction \cite{actionconditional, videoprediction} as an example, the loss function is usually future frame reconstruction error. One failure case is that these methods may waste effort reconstructing non-generalizable information, such as textures or the background. This is analogous to the concept of \textit{robust} and \textit{non-robust} features from supervised learning \cite{robustness_accuracy, adv_bugs}. In RL, \textit{non-robust} SL features are analogous to \textit{non-generalizable} RL features across the MDP family. If the dynamics model provides non-generalizable features, it can improve sample efficiency for optimizing the training MDP's, analogous to the SL setting where providing non-robust features improves an SL classifier's accuracy, but at the expense of robustness (in SL) or generalization (in RL).

Many of these effects can be simulated by modifying the observation function ${w_{\theta}: \mathcal{S} \rightarrow \mathcal{O}}$ that takes states to observations. As an example, the low-dimensional representation of $xy$-coordinates of a character in a game may be projected to a large image in observation space. Adding non-generalizable features which are still correlated with intra-MDP progress \cite{visualizing} can cause an agent to overfit. For generalization, a dynamics model should capture the dynamics in the state space, rather than the observation space.
Depending on the MDP, modelling much of the dynamics
may be necessary to learn the optimal action. If the dynamics function $\mathcal{T}$ is difficult to predict, the generalization bottleneck comes primarily from navigating those hard-to-predict dynamics, possibly with assistance from a dynamics model.
Using ideas from cryptography, in Appendix \ref{appendix} we construct extreme examples of observation functions and model dynamics, where we show that generalization can be at odds with good sample complexity in model-based RL.

\section{Conclusion}
We have stated that an important condition for RL generalization is for the benchmark to possess the principle of unchanged optimality. This leads to the requirement that the benchmark should therefore have some way of producing information necessary to output the optimal action. We have also discussed the ways which this affects model-based RL techniques. We hope that this work provides future direction through the subtleties and complications of RL generalization.

\newpage 
\bibliography{references}
\bibliographystyle{icml2019}

\input{appendix.tex}

\end{document}

%% file: appendix.tex
\onecolumn
\appendix
\renewcommand{\thesection}{A.\arabic{section}}
\renewcommand{\thefigure}{A\arabic{figure}}
\setcounter{section}{0}
\setcounter{figure}{0}
\title{Appendix: The Principle of Unchanged Optimality in Reinforcement Learning Generalization}
\date{}
\maketitle

\section{Extreme Examples for Model Based RL Generalization} \label{appendix}
In order to draw instructive and extreme examples to show cases of RL generalization, we draw analogies from cryptography. When the observation may obfuscate important information and becomes hard to invert, this is similar to a \textit{hash} function \cite{crypto}. Similarly, if a dynamics is difficult to predict, this is similar to a \textit{pseudorandom} function.

As a brief overview to the reader, we use functions inherently requiring a large sample efficiency to break \cite{crypto}, but we do not use the fully formal definitions of indistinguishability in cryptography, for the sake of clarity. However, our examples are still easily modifiable to the official definitions if needed. A \textit{one-way} function family, or a hash function family, is a set of functions $f_{\phi}: \{0, 1\}^{m} \rightarrow \{0,1\}^{m} $ such that it requires more than $poly(m)$ queries to invert for more than $poly(m)$ outputs, while a \textit{pseudorandom} function family $g_{\phi}: \{ 0,1\}^{m} \rightarrow \{0,1\}^{m}$ is a set of functions such that it takes more than $poly(m)$ queries to predict the output on more than $poly(m)$ inputs.

\subsection{Obfuscating Observation Functions}
We show that the sample complexity of generalization can be independent to predicting the observational dynamics and the reward dynamics. 

Let $\mathcal{H}_{m} = \{0,1\}^{m}$ be the boolean hypercube set. Suppose all parts of our MDP family were fixed and relatively simple functions except the observation function. 

\begin{itemize}
\item For a seed $\theta$, we let reward $r_{\theta}(s) = 1$ if $s = c_{\theta}$ and $0$ otherwise, and we will allow the policy to know $c_{\theta}$ at the start of the MDP. 

\item We let the observation function be a bijective hash function, i.e. $w: \mathcal{S} \rightarrow \mathcal{O}$ whose output is what the policy actually sees. By definition of the hash function, the sample complexity of $r$ will be more than $poly(m)$. 

\item Let the transition on the observation space be $\mathcal{T}_{obs}(o, a) = o + a$, which is the boolean addition of observation with action, which means the underlying transition will be $\mathcal{T}(s, a) = w^{-1} \circ \mathcal{T}_{obs}( w(s), a)$. Let the time horizon be $T = poly(m)$. 

\item Lastly, let $s_{0}$ be chosen uniformly random from $\mathcal{H}_{m}$ using seed $\theta$.
\end{itemize}

Then we define the family of MDP's as $\mathcal{M}_{\theta} = (\mathcal{S} = \mathcal{H}_{m}, \mathcal{A} = \mathcal{H}_{m},\mathcal{O} = \mathcal{H}_{m}, \mathcal{S}_{0} = Rand_{\theta}( \mathcal{H}_{m}),   r_{\theta}, \mathcal{T}, w)$.

Note that however, it is easy for a dynamics model to predict $\mathcal{T}_{obs}$ with very few $poly(m)$ samples. 

Furthermore, an overfitting policy may memorize trajectories - i.e. for each seed $\theta$ it may observe starting position $w(s^{\theta}_{0})$ and thus it may, through random search, find the optimal action sequence $a^{\theta}_{0},...,a^{\theta}_{t},...$ associated with the initial starting observation in order to maximize reward. The policy may not know which states it is on (and thus how the reward function works), but it still may observe $w(s^{\theta}_{t})$ which will allow it to output the next optimal action if it has memorized the trajectory associated with $\theta$.

The idea here is that in order for a policy to generalize, it must figure out how to invert $w$, so that the policy has control and knowledge over its positions in the underlying state space $\mathcal{S}$. It will solve $w^{-1}$ by generating pairs $(o, r_{\theta} \circ w^{-1}(o))$ at training time, to eventually figure out $w^{-1}$. Note however, the optimal policy will need more than $poly(m)$ such pairs to do so, and thus requires very high sample complexity.

Furthermore, \textit{any} policy which optimizes all MDP's in the family must possess the invertibility condition. Achieving maximal reward implies that $ s_{1} = c_{\theta}$, and thus the policy chose $a_{0}$ such that $  \mathcal{T}(s_{0}, a_{0}) = c_{\theta} \iff w^{-1} (w(s_{0}) + a_{0}) = c_{\theta}$. As the seed $\theta$ ranges, this implies that the policy knows the correct action $a$ for any pair $(s,c)$ satisfying the mentioned condition, implying invertibility power.

\subsection{Pseudorandom Dynamics}
We show that the sample complexity of generalization can be \textit{nearly equivalent} to predicting the dynamics.

\begin{itemize}
\item For a seed $\theta$, we let reward $r_{\theta}(s) = 1$ if $s = c_{\theta}$ and $0$ otherwise, and we will allow the policy to know $c_{\theta}$ at the start of the MDP. 

\item Suppose $\mathcal{T}$ was a pseudo-random function, and our time horizon was $T = poly(m)$.

\item Lastly, let $s_{0}$ be chosen uniformly random from $\mathcal{H}_{m}$ using seed $\theta$.
\end{itemize}

Thus we have $\mathcal{M}_{\theta} = (\mathcal{S} = \mathcal{H}_{m}, \mathcal{A} = \mathcal{H}_{m}, r_{\theta}, \mathcal{T}, \mathcal{S}_{0} = Rand_{\theta}( \mathcal{H}_{m}) )$.

An overfitting policy for a given $\theta$ using random search may eventually find the optimal state-action sequence $\mathbf{s}_{\theta} = (s_{0}, a_{constant}, s_{1}, a_{1},...,)$ which nearly maximizes the reward $r_{\theta}$ - thus in order to optimize on multiple $\{\theta_{1},...,\theta_{n}\}$, the policy may then use $(s_{0},a_{constant})$ as an anchor and observe $s_{1}$ to know which $\theta_{i}$ it is on and recall from memory, the optimal path $\mathbf{s}_{\theta_{i}}$. 

The idea here is that the policy is once again, unable to control its trajectory in the state space unless it understands how to predict $\mathcal{T}$. An example of an optimal policy may at training time, query an exponentially large number of pairs $(s_{t}, a_{t}, s_{t+1})$ in order to produce a dynamics model $\mathcal{T}$, from which the policy may then have full control over its state trajectories and be able to, at test time, optimize $r_{\theta}$. 

Any optimal policy will immediately jump to the correct state $c_{\theta}$ to optimize the reward, and thus satisfies $\mathcal{T}(s^{\theta}_{0}, a_{0}) = c_{\theta} $. As $\theta$ ranges across all seeds, this implies that the policy knows the correct $a$ for all $(s, c)$ pairs such that $\mathcal{T}(s, a) = c$, implying predictive power on $\mathcal{T}$.